\documentclass[conference]{IEEEtran}
\IEEEoverridecommandlockouts
\usepackage{cite}
\usepackage{amsmath,amssymb,amsfonts}
\usepackage{algorithmic}
\usepackage{graphicx}
\usepackage{textcomp}
\usepackage{xcolor}
\usepackage{booktabs}
\usepackage{makecell}
\usepackage[linesnumbered,ruled,vlined]{algorithm2e}
\SetKwInput{KwInput}{Input}
\SetKwInput{KwOutput}

\usepackage{url}

\usepackage{tikz}
\usetikzlibrary{arrows.meta, positioning, shapes.geometric, shapes.symbols, fit}

\def\BibTeX{{\rm B\kern-.05em{\sc i\kern-.025em b}\kern-.08em
    T\kern-.1667em\lower.7ex\hbox{E}\kern-.125emX}}
\DeclareUnicodeCharacter{2208}{\ensuremath{\in}}
\begin{document}

\title{Physically Constrained Kinematic Calibration for High-Precision Robotic Needle Guidance\\
}
\author{
    \IEEEauthorblockN{
        Prabhu Nithin Gollapudi\IEEEauthorrefmark{1},
        Andreas Maier\IEEEauthorrefmark{1},
        Franziska Mathis-Ullrich\IEEEauthorrefmark{1},
        Water Zhao\IEEEauthorrefmark{2},
        Jia Wang\IEEEauthorrefmark{2}, and
        Siming Bayer\IEEEauthorrefmark{1}
    }
    \IEEEauthorblockA{
        \IEEEauthorrefmark{1} Friedrich-Alexander-Universität Erlangen-Nürnberg, Erlangen, Germany\\
        Email: \{prabhu.nithin.gollapudi, andreas.maier, franziska.mathis-ullrich, siming.bayer\}@fau.de
    }
    \IEEEauthorblockA{
        \IEEEauthorrefmark{2}Precision Medtech Co., Ltd, Beijing, China\\
        Email: \{water.zhao, jia.wang\}@precisionmedtech.com
    }
}

\maketitle

\begin{abstract}

Percutaneous needle insertion is a common practice in minimally invasive procedures whose clinical success depends on accurate needle placement. However, achieving the necessary precision is a challenge for manual operators. Robotic assistance can improve targeting accuracy and reduce operator dependence, but residual millimeter-scale errors persist because nominal kinematic models do not fully represent the physical manipulator. This paper presents a robust bound-constrained kinematic calibration method designed to improve positional accuracy for a 6-Degree-of-Freedom (6-DoF) robotic needle-guidance manipulator. The approach uses physically plausible parameter ranges to regularize a compact 19-parameter geometric identification model. By treating these bounds as active optimization constraints rather than mere numerical safeguards, the framework effectively mitigates parameter coupling and overfitting, ensuring highly stable parameter identification. In simulation, the proposed method reduced end-effector Root Mean Square Error (RMSE) from 3.422\,mm to 0.690\,mm (a 79.8\% improvement). In benchtop hardware experiments, the calibrated model achieved sub-millimeter validation accuracy, including 0.865\,mm RMSE in experimental trajectory validation and 0.610\,mm RMSE in a coordinate-lattice verification task. These results demonstrate that physically constrained kinematic calibration can enhance robotic needle-targeting accuracy, establishing a high-precision foundation for future clinical applications.

\end{abstract}


\section{Introduction}

Accurate percutaneous needle insertion is a critical component of various minimally invasive interventions, including biopsies, localized drug delivery, and thermal ablation. Among these, percutaneous thermal ablation for liver tumors represents a particularly demanding application. While ablation serves as a primary therapeutic alternative for patients ineligible for surgical resection \cite{globocan, liang, uribe}, its clinical efficacy relies heavily on precise targeting. Achieving complete tumor necrosis requires a strict 5--10\,mm safety margin while carefully avoiding collateral damage to adjacent vasculature and bile ducts \cite{tinguely, beyer, Schaible2020-go, bodard}. Even small targeting deviations can lead to incomplete ablation or severe complications. Consequently, robotic assistance has emerged as a promising solution for reducing operator variability and enabling consistent, high-precision workflows \cite{uribe,bodard,mbalisike}.

Despite these advantages, the absolute positioning accuracy of current robotic systems is still limited by discrepancies between the nominal kinematic model provided by the manufacturer and the actual physical manipulator. These discrepancies, caused by manufacturing tolerances, assembly errors, joint compliance, and encoder offsets, typically result in millimeter-scale Cartesian positioning errors at the end-effector~\cite{gao,roth}.

Kinematic calibration is the established method for compensating for such geometric inaccuracies in serial manipulators. However, existing approaches often face a fundamental trade-off between model fidelity and robustness. High-dimensional models suffer from parameter coupling and overfitting, while simplified models may miss critical error sources. Moreover, most studies emphasize in-sample fitting accuracy with limited attention to parameter stability, generalization performance, or the incorporation of physical constraints~\cite{NUBIOLA2013236, gao,7759386}.

To address these limitations, this paper proposes a robust kinematic calibration framework for a 6-DoF serial manipulator tailored for percutaneous needle guidance. Rather than treating parameter bounds merely as numerical safeguards, our approach utilizes physically motivated bounds as an explicit regularization mechanism to ensure parameter stability and predictive accuracy.

The specific contributions of this work are:
\begin{itemize}
\item A 19-parameter geometric identification model that captures the dominant error sources of the manipulator.
\item A calibration pipeline combining observability-optimized configuration sampling, Kabsch-based tracker-to-base initialization, and bound-constrained Trust-Region Reflective (TRF) optimization.
\item Comprehensive benchtop validation utilizing a 10-fold cross-validation methodology, simulation, and physical hardware experiments that demonstrates robust sub-millimeter positioning accuracy. This establishes the fundamental kinematic precision required prior to future phantom and \textit{ex vivo} studies.
\end{itemize}

The remainder of this paper is organized as follows: Section II reviews related work. Section III details the kinematic modeling and the proposed calibration methodology. Section IV describes the experiments and results, and Section V concludes the paper.

\section{Related Work}

Kinematic calibration is essential for achieving the absolute positioning accuracy required in robot-assisted percutaneous interventions. In demanding applications such as liver tumor ablation, even small targeting errors may compromise tumor coverage or damage nearby vasculature~\cite{uribe,bodard}. Nominal manufacturer models typically neglect manufacturing tolerances, assembly errors, and joint encoder offsets, which can lead to millimeter-scale geometric inaccuracies in practice~\cite{klimchik2012, gao,roth}. In serial manipulators, these discrepancies are commonly modeled using either the Denavit-Hartenberg (DH) convention or the Product of Exponentials (PoE) formulation, while parameter identification is typically posed as a nonlinear least-squares problem and solved using gradient-based or population-based optimization methods~\cite{he,gao,cui, klimchik2012, NUBIOLA2013236}. To improve numerical conditioning and identifiability, many studies further incorporate observability analysis or singular value decomposition to guide pose selection and reduce parameter redundancy~\cite{wu13, klimchik2012, gao,7759386}.

In medical and interventional robotics, calibration must balance model expressiveness with practical constraints such as limited workspace, controlled measurement conditions, and fast experimental deployment. Prior studies have reported substantial accuracy gains after calibration. For example, Yunfei \textit{et al.} demonstrated that a compact calibration model for 6R serial robots achieved strong improvements in both simulation and hardware experiments~\cite{20150412}. Similarly, Cui \textit{et al.} reduced mean positioning error from 1.11\,mm to 0.17\,mm for a 6-DoF industrial robot~\cite{cui}, and Kana \textit{et al.} reported a reduction from 12.5\,mm to 5.8\,mm using a fast recalibration strategy~\cite{kana}. These results confirm the practical value of calibration for robotic targeting tasks~\cite{bodard,uribe,mbalisike}.

Despite these advances, important limitations remain. High-parameter models can suffer from parameter coupling and sensitivity to the selected calibration poses, whereas overly simplified models may fail to capture the dominant geometric error sources adequately~\cite{7759386,gao}. In addition, many studies focus primarily on fit quality under a single dataset or experimental setup, with limited emphasis on parameter stability, out-of-sample validation, or physically constrained identification. These issues are particularly relevant in interventional robotics, where calibration should remain robust across datasets. Motivated by these gaps, this work proposes a compact 19-parameter geometric calibration framework that combines observability-aware sampling, physically motivated parameter bounds, and 10-fold cross-validation to improve identification robustness and generalization.

\section{Methodology}

The robotic platform utilized in this study is a proprietary 6-DoF serial manipulator provided by Precision Medtech Co., Ltd. Due to confidentiality agreements and intellectual property restrictions, the full nominal DH parameter table cannot be disclosed.

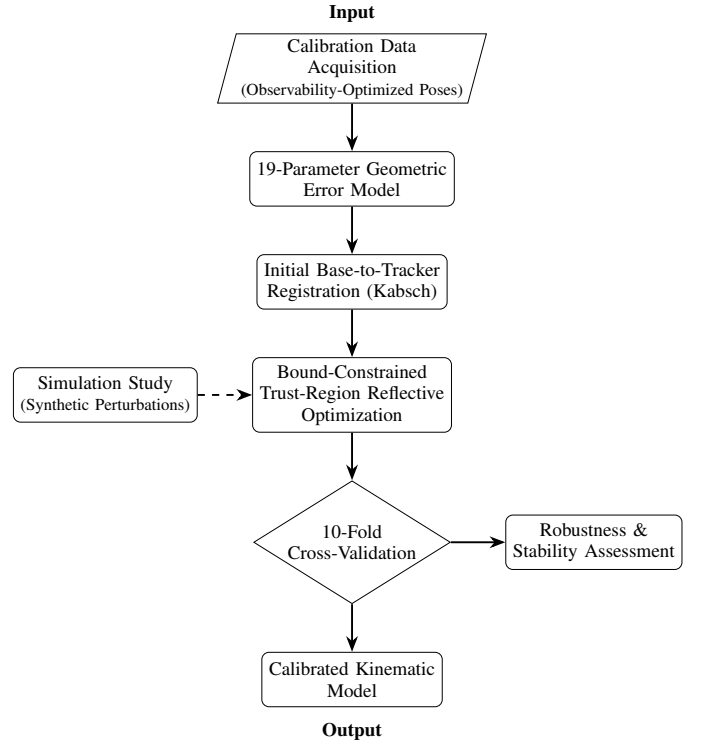
\begin{figure}[!t]
\centering
\resizebox{\columnwidth}{!}{%
\begin{tikzpicture}[
    node distance=0.7cm and 0.8cm,
    font=\footnotesize,
    box/.style={rectangle, draw, rounded corners=3pt, minimum width=2.4cm, minimum height=0.8cm, align=center, inner sep=3pt},
    arrow/.style={-{Stealth}, thick},
    process/.style={box},
    decision/.style={diamond, draw, aspect=1.6, minimum width=2.4cm, minimum height=0.8cm, align=center, inner sep=1pt},
    data/.style={trapezium, trapezium left angle=75, trapezium right angle=105, draw, minimum width=2.4cm, minimum height=0.8cm, align=center, inner sep=2pt}
]

\node[data] (data) {Calibration Data\\Acquisition\\{\scriptsize(Observability-Optimized Poses)}};
\node[process, below=of data] (model) {19-Parameter Geometric\\Error Model};
\node[process, below=of model] (kabsch) {Initial Base-to-Tracker\\Registration (Kabsch)};
\node[process, below=of kabsch] (opt) {Bound-Constrained\\Trust-Region Reflective\\Optimization};
\node[decision, below=of opt] (cv) {10-Fold\\Cross-Validation};
\node[process, right=of cv] (eval) {Robustness \&\\Stability Assessment};
\node[process, below=of cv] (output) {Calibrated Kinematic\\Model};

\node[process, left=of opt] (sim) {Simulation Study\\{\scriptsize(Synthetic Perturbations)}};

\draw[arrow] (data) -- (model);
\draw[arrow] (model) -- (kabsch);
\draw[arrow] (kabsch) -- (opt);
\draw[arrow] (opt) -- (cv);
\draw[arrow] (cv) -- (output);

\draw[arrow] (cv) -- (eval);
\draw[arrow, dashed] (sim) -- (opt); 

\node[above=0.05cm of data, font=\footnotesize\bfseries] {Input};
\node[below=0.1cm of output, font=\footnotesize\bfseries] {Output};

\end{tikzpicture}%
}
\caption{Calibration pipeline overview. The framework combines observability-optimized data acquisition, a compact 19-parameter error model, bound-constrained optimization, and 10-fold cross-validation for robust parameter identification. The simulation branch (dashed) is used for controlled validation before hardware experiments.}
\label{fig:calibration_pipeline}
\end{figure}

Because the underlying physical architecture follows a standard 6-DoF serial morphology, the baseline Forward Kinematics (FK) and Jacobian computations are derived using the conventional DH convention~\cite{10.1115/1.4011045, spong}. Rather than reproducing these standard textbook formulations, this section focuses explicitly on the novel aspects of our framework: the observability-optimized data acquisition, the formulation of the reduced 19-parameter error model, and the bound-constrained optimization pipeline (illustrated in Fig.~\ref{fig:calibration_pipeline}).

\subsection{Calibration Data Acquisition}
\label{subsec:data_acquisition}

Calibration poses were selected from an observability-optimized set of joint configurations to reduce parameter coupling and improve identification robustness~\cite{4543308, 7759386, klimchik2012}. For each sampled pose $k \in \{1,\dots,N\}$, the robot was commanded to

\begin{equation}
\mathbf{q}^{(k)} = \left[\theta_1^{(k)}, \theta_2^{(k)}, \dots, \theta_6^{(k)}\right]^T.
\end{equation}

After the robot settles, two Cartesian position vectors are associated with the same pose: the nominal end-effector position $\mathbf{p}_{\mathrm{nom}}^{(k)} \in \mathbb{R}^3$, computed using the manufacturer's DH parameters, and the measured position $\mathbf{p}_{\mathrm{meas}}^{(k)} \in \mathbb{R}^3$, obtained using a Leica AT960 laser tracker observing a spherically mounted retroreflector (SMR) attached to the end-effector. The resulting calibration dataset is
\begin{equation}
\mathcal{D}_{\mathrm{cal}} =
\left\{
\left(
\mathbf{q}^{(k)},
\mathbf{p}_{\mathrm{nom}}^{(k)},
\mathbf{p}_{\mathrm{meas}}^{(k)}
\right)
\right\}_{k=1}^{N}.
\label{eq:dcal}
\end{equation}

\subsection{Reduced 19-Parameter Calibration Model}
\label{subsec:19param_model}

A full kinematic model contains 24 internal DH parameters. However, many of these exhibit strong coupling and low observability, leading to overfitting. Furthermore, the manufacturer specifically required that only a restricted subset of parameters be updated in the robot's controller. To accommodate these practical constraints, we adopt a reduced model consisting of 13 internal parameters (seven geometric length corrections and six joint angle offsets) alongside six external parameters describing the base-to-tracker rigid transformation.

Table~\ref{tab:dh_error_mapping} details the exact distribution of the 13 internal structural variations relative to their respective links. Notably, by keeping link twist errors fixed ($\Delta \alpha_i = 0$), the kinematic model remains safe from the parameter singularities that typically occur in the conventional DH representation when consecutive joint axes are parallel or near-parallel (such as between Joint~2 and Joint~3 on this platform)~\cite{https://doi.org/10.1002/rob.4620020406, mooring}. This targeted parameter omission eliminates the need for complex structural model modifications.

\begin{table}[ht]
\centering
\caption{Mapping of the 13 Kinematic Error Parameters to the DH Table}
\label{tab:dh_error_mapping}
\begin{tabular}{@{}lcccc@{}}
\toprule
Link $i$ & $\Delta a_{i}$ (mm) & $\Delta \alpha_{i}$ (rad) & $ \Delta d_i$ (mm) & $\Delta  \theta_i$ (rad) \\ \midrule
1 & $\Delta x_1$ & 0 & $\Delta z_1$ & $\Delta \theta_1$ \\
2 & $\Delta a$   & 0 & 0            & $\Delta \theta_2$ \\
3 & $\Delta y_3$ & 0 & 0            & $\Delta \theta_3$ \\
4 & 0            & 0 & $\Delta z_4$ & $\Delta \theta_4$ \\
5 & 0            & 0 & 0            & $\Delta \theta_5$ \\
6 & $\Delta y_6$ & 0 & $\Delta z_6$ & $\Delta \theta_6$ \\ \bottomrule
\end{tabular}
\end{table}

The internal geometric and joint angle parameters are defined as
\begin{equation}
\Delta \boldsymbol{\phi}_{\mathrm{geom}} =
\left[
\Delta x_1,\,
\Delta z_1,\,
\Delta a,\,
\Delta y_3,\,
\Delta z_4,\,
\Delta y_6,\,
\Delta z_6
\right]^T,
\label{eq:geom_params}
\end{equation}
\begin{equation}
\Delta \boldsymbol{\phi}_{\mathrm{angle}} =
\left[
\Delta \theta_1,\,
\dots,
\Delta \theta_6
\right]^T.
\label{eq:angle_params}
\end{equation}

The six external parameters describing the base-to-tracker transformation consist of a rotational component and a translational component:
\begin{equation}
\Delta \boldsymbol{\phi}_{\mathrm{ext}} =
\left[
r_x,\,
r_y,\,
r_z,\,
t_x,\,
t_y,\,
t_z
\right]^T,
\label{eq:external_params}
\end{equation}
yielding the full calibration vector
\begin{equation}
\Delta \boldsymbol{\phi} =
\left[
\Delta \boldsymbol{\phi}_{\mathrm{geom}}^T \;
\Delta \boldsymbol{\phi}_{\mathrm{angle}}^T \;
\Delta \boldsymbol{\phi}_{\mathrm{ext}}^T
\right]^T \in \mathbb{R}^{19},
\label{eq:full_param_vector}
\end{equation}
where the rotational components ($r_x, r_y, r_z$) are parameterized as a compact rotation vector (axis-angle representation), whose direction defines the axis of rotation and whose Euclidean norm defines the angle of rotation in radians.

A major advantage of coupling this specific parameter selection with the bound-constrained optimization proposed in Section~\ref{subsec:bounded_identification} is its high adaptability. The framework allows for the identification of any arbitrary subset of DH parameters simply by adjusting the optimization limits ($\mathbf{l}$ and $\mathbf{u}$). Parameters restricted by manufacturer specifications can be effectively locked by setting their upper and lower bounds to zero, seamlessly accommodating industrial constraints without requiring modifications to the underlying Jacobian or error functions.

\subsection{Bound-Constrained Parameter Identification}
\label{subsec:bounded_identification}

An initial estimate of the base-to-tracker transformation ($\Delta \boldsymbol{\phi}_{\mathrm{ext}}$) is obtained via the Kabsch algorithm \cite{https://doi.org/10.1107/S0567739476001873}. It is subsequently refined jointly with the internal parameters. The calibration solves the following bounded nonlinear least-squares problem:
\begin{equation}
\min_{\Delta \boldsymbol{\phi}}
\frac{1}{2}
\sum_{k=1}^{N}
\left\|
\mathbf{p}_{\mathrm{meas}}^{(k)} -
\hat{\mathbf{p}}^{(k)}(\mathbf{q}^{(k)}, \Delta \boldsymbol{\phi})
\right\|_2^2
\label{eq:bounded_obj}
\end{equation}
subject to
\begin{equation}
\mathbf{l} \leq \Delta \boldsymbol{\phi} \leq \mathbf{u},
\label{eq:bounded_constraints}
\end{equation}
where $\mathbf{l}$ and $\mathbf{u}$ denote the lower and upper parameter bounds, respectively, and $\hat{\mathbf{p}}^{(k)}$ is the $3 \times 1$ Cartesian position vector extracted from the homogeneous transformation matrix produced by the forward kinematics model at joint configuration $\mathbf{q}^{(k)}$.

By treating bounds as an explicit regularization mechanism rather than merely numerical safeguards, the approach improves identification stability and reduces the risk of physically implausible solutions.

Prior to optimization, the internal geometric and joint angle parameters ($\Delta \boldsymbol{\phi}_{\mathrm{geom}}$ and $\Delta \boldsymbol{\phi}_{\mathrm{angle}}$) are initialized to zero. To ensure rapid and stable convergence, the Trust-Region Reflective algorithm is supplied with an analytically derived Jacobian matrix. The optimization stopping criteria are defined by a strictly enforced function tolerance of $10^{-9}$, a step tolerance of $10^{-9}$, and a maximum limit of $5000$ iterations.

\subsection{Robustness Assessment via $K$-Fold Cross-Validation}
\label{subsec:kfold}

To quantify generalization and parameter stability, the calibration is embedded in a 10-fold cross-validation framework. For each fold $i$, parameters $  \Delta \boldsymbol{\phi}^{(i)}  $ are identified on the training subset and evaluated on the held-out validation data. This yields fold-wise error statistics and parameter variance across folds. A robust calibration solution is characterized by low out-of-sample RMSE combined with low inter-fold parameter variability.

\subsection{Simulation Protocol}
\label{subsec:simulation_protocol}

A controlled simulation environment was developed to validate convergence and sensitivity prior to hardware experiments following literature~\cite{20150412, 9791336}. Synthetic datasets were generated by injecting known geometric perturbations, encoder biases, and realistic measurement noise into the nominal forward kinematics model. This enabled direct assessment of parameter recovery accuracy and the effect of bound tightness.

\section{Experiments and Results}
\label{sec:experiments_results}

This section evaluates the proposed 19-parameter bound-constrained calibration framework through simulation and physical experiments on a 6-DoF serial manipulator. The evaluation focuses on Cartesian positioning accuracy and the robustness of the identified parameters on unseen data.

\subsection{Evaluation Metrics}
\label{subsec:eval_metrics}

To ensure unambiguous comparison with existing literature, performance is quantified using the 3D Euclidean distance between the laser-tracker ground truth and the forward-kinematics predictions, computed strictly within the laser tracker coordinate frame. For a dataset of $N_e$ evaluation poses, the Root Mean Square Error (RMSE) is defined as:
\begin{equation}
E_{\mathrm{RMSE}} = \sqrt{\frac{1}{N_e} \sum_{i=1}^{N_e} \left\| \mathbf{p}_{\mathrm{meas}}^{(i)} - \hat{\mathbf{p}}_{\mathrm{laser}}^{(i)} \right\|_2^2},
\label{eq:rmse}
\end{equation}
where $\mathbf{p}_{\mathrm{meas}}^{(i)}$ is the measured spatial coordinate and $\hat{\mathbf{p}}_{\mathrm{laser}}^{(i)}$ is the model prediction transformed into the laser tracker frame using the identified external parameters ($\Delta \boldsymbol{\phi}_{\mathrm{ext}}$). 

Mean and maximum absolute errors are also reported to characterize average and worst-case performance.

\subsection{Experimental Setup}
\label{subsec:exp_setup}

All physical measurements were performed with a Leica Absolute Tracker AT960 and an SMR mounted on the end-effector. The same measurement protocol was used for both nominal and calibrated evaluations to ensure a fair comparison.

\subsection{Simulation Results}
\label{subsec:sim_results}

A controlled simulation study with $  N=120  $ observability-optimized configurations was conducted to assess parameter recovery under known perturbations (geometric, encoder, and measurement noise). The dataset was split evenly into training and testing sets. Table~\ref{tab:sim_results} summarizes the resulting Cartesian errors before and after calibration. The proposed method reduced the testing RMSE from 3.422 mm to 0.690 mm, corresponding to a 79.8\% reduction, while also reducing both the mean and maximum residual errors.

\begin{table}[t]
\centering
\caption{Simulation results before and after calibration.}
\label{tab:sim_results}
\begin{tabular}{lccc}
\toprule
Configuration State & RMSE (mm) & Mean (mm) & Max (mm) \\
\midrule
Nominal (train) & 2.645 & 2.402 & 4.800 \\
Nominal (test)  & 3.422 & 3.010 & 7.662 \\
Calibrated (train)  & 0.578 & 0.516 & 1.411 \\
Calibrated (test)   & 0.690 & 0.609 & 1.258 \\
\bottomrule
\end{tabular}
\end{table}

\subsection{Offline Physical Validation}
\label{subsec:offline_validation}

The physical calibration framework was evaluated using three experimentally acquired datasets with different sampling profiles. Dataset A contained 136 observability-optimized samples, Dataset B contained 60 randomly sampled configurations, and Dataset C contained 52 samples with deliberately introduced joint-angle offsets to test compensation under larger structural deviations.

\begin{figure}[t]
    \centering
    \includegraphics[width=\columnwidth]{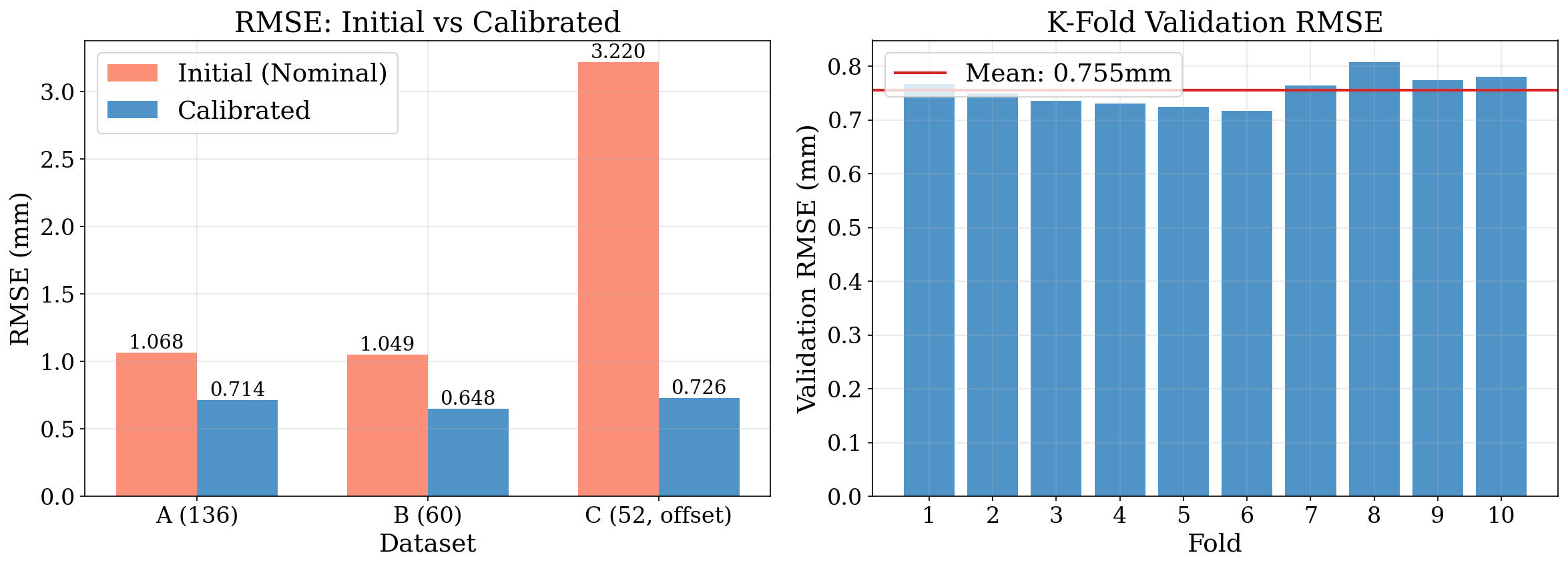}
    \caption{Validation RMSE across the 10 folds for the nominal and calibrated models. The proposed framework consistently reduces error and exhibits high robustness.}
    \label{fig:cv_rmse}
\end{figure}

The pooled dataset was evaluated using 10-fold cross-validation with physically motivated bounds of $\pm 3.0$\,mm for geometric parameters and $\pm 2.0^\circ$ for angular parameters. Figure~\ref{fig:cv_rmse} shows the validation RMSE across folds. The calibrated model consistently achieved lower errors with low fold-to-fold variability, yielding a pooled validation RMSE of 0.755\,mm.

\begin{figure}[ht]
\centering
\includegraphics[width=\columnwidth]{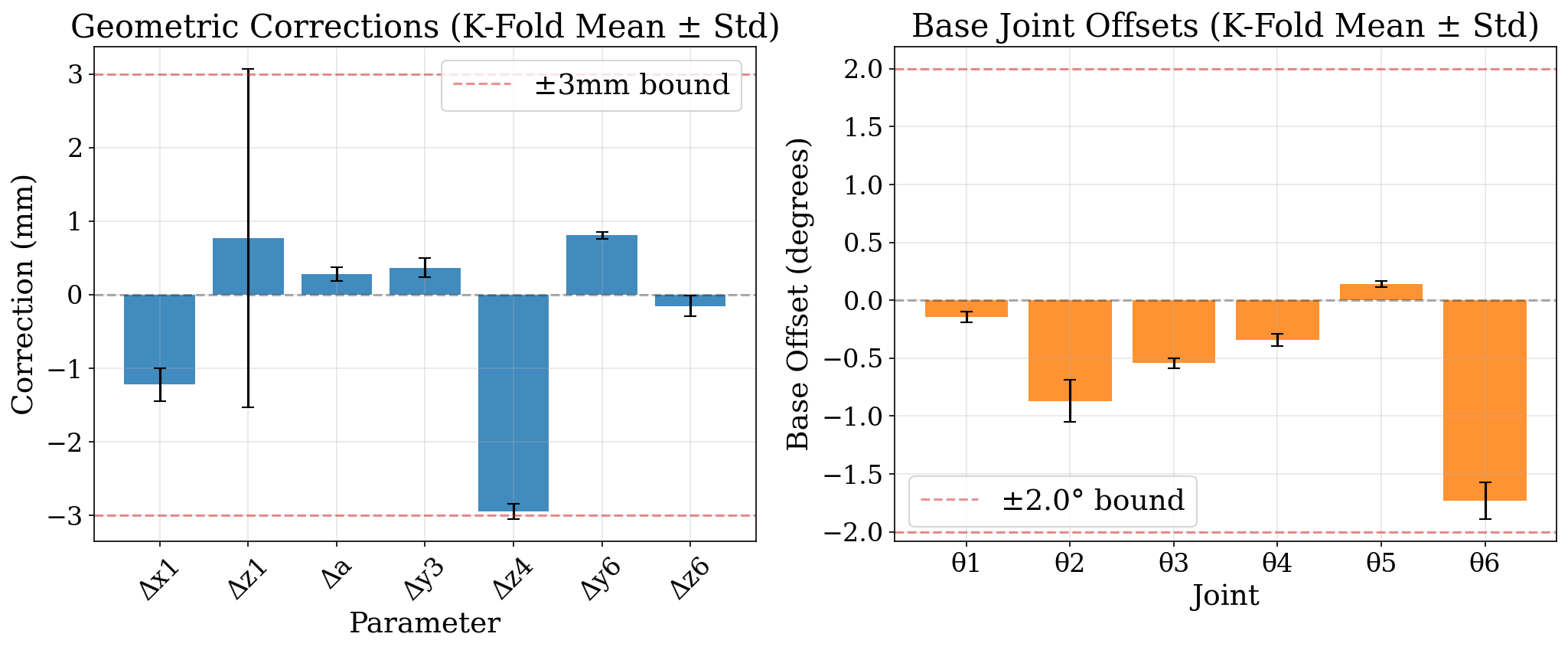}
\caption{Estimated kinematic error parameters tracked across the ten independent validation folds. The low fold-to-fold variance demonstrates the high identifiability and robustness achieved by the bound-constrained optimization.}
\label{fig:calibration_values}
\end{figure}

To further verify that the bound-constrained optimization successfully prevented overfitting, the numerical stability of the identified parameters was analyzed. Figure~\ref{fig:calibration_values} illustrates the variation of the 19 kinematic parameters across each of the ten independent cross-validation folds. The narrow variance profile confirms that the parameters converge to a consistent, physically meaningful minimum regardless of the specific data subset. This stability establishes the baseline for the final mean parameter values applied to the robot controller.

Table~\ref{tab:offline_results} summarizes per-dataset performance. The largest improvement occurred on Dataset C (RMSE reduced from 3.220\,mm to 0.726\,mm), demonstrating effective compensation even under larger structural deviations.

\begin{table}[t]
\centering
\caption{Offline physical calibration results.}
\label{tab:offline_results}
\begin{tabular}{lccc}
\toprule
Dataset / State & RMSE (mm) & Mean (mm) & Max (mm) \\
\midrule
A (nominal) & 1.068 & 0.972 & 2.214 \\
A (calibrated)  & 0.714 & 0.644 & 2.000 \\
\midrule
B (nominal) & 1.049 & 0.964 & 1.880 \\
B (calibrated)   & 0.648 & 0.585 & 1.801 \\
\midrule
C (nominal) & 3.220 & 3.050 & 5.008 \\
C (calibrated)   & 0.726 & 0.668 & 1.442 \\
\bottomrule
\end{tabular}
\end{table}

\subsection{Independent Experimental Validation}
\label{subsec:online_validation}

To verify out-of-sample performance, the final calibrated model was deployed on the physical robot and tested on an independent trajectory of 60 previously unseen configurations. As reported in Table~\ref{tab:online_results}, the validation RMSE improved from 1.049\,mm (nominal) to 0.865\,mm (calibrated).

\begin{table}[t]
\centering
\caption{Independent experimental validation on the physical robot.}
\label{tab:online_results}
\begin{tabular}{lccc}
\toprule
Model State & RMSE (mm) & Mean (mm) & Max (mm) \\
\midrule
Nominal model    & 1.049 & 0.964 & 1.837 \\
Calibrated model & 0.865 & 0.804 & 1.598 \\
\bottomrule
\end{tabular}
\end{table}

\begin{figure}[ht]
\centering
\includegraphics[width=\columnwidth]{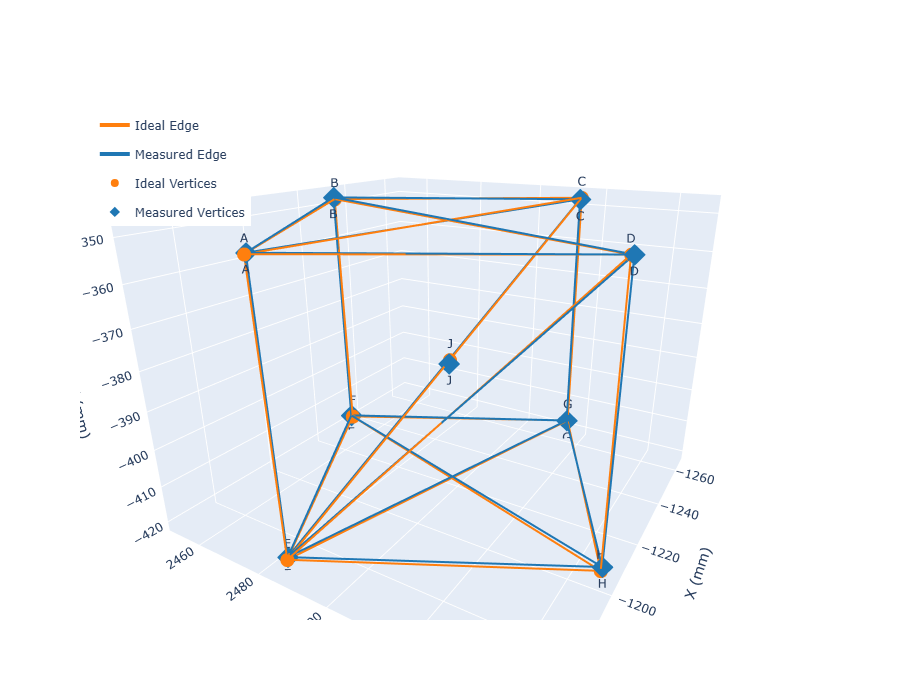}
\caption{Three-dimensional coordinate lattice test after calibration, illustrating the spatial alignment between commanded target positions and actual measured positions.}
\label{fig:lattice_test_result}
\end{figure}

As a complementary spatial verification step, a physical coordinate lattice test was executed by commanding the manipulator across a distinct grid of target positions forming a cuboidal workspace envelope. Figure~\ref{fig:lattice_test_result} illustrates the spatial mapping between the commanded coordinate grid points and the actual measured tracker positions using the calibrated model. This structured test visually confirms the robust sub-millimeter targeting performance across the operational workspace, yielding a final RMSE of 0.610\,mm.

\subsection{Discussion}
\label{subsec:robustness_discussion}

The proposed 19-parameter framework successfully suppresses out-of-sample positioning error, delivering a accuracy gain over typical uncompensated serial manipulators (which often exhibit 2.00--5.00\,mm errors~\cite{gao}). By consistently achieving sub-millimeter precision, the calibrated model fulfills the strict engineering tolerances required for future percutaneous thermal ablation applications. The modest residual gap between the physical hardware validation (0.865\,mm) and the simulated baseline (0.690\,mm) is primarily attributable to unmodeled non-geometric effects, such as joint compliance, gravitational deflection, and optical tracking jitter. Furthermore, the lower error observed in the coordinate lattice test (0.610\,mm) relative to the random trajectory is consistent with its geometrically restricted envelope, which limits exposure to workspace boundaries where residual compliance effects are most pronounced.

The cross-validation analysis of the identified parameters (Fig.~\ref{fig:calibration_values}) highlights the critical importance of the bound-constrained optimization strategy. While most parameters converged with high stability, isolated parameters (such as $\Delta z_4$ and $\Delta \theta_6$) occasionally saturated near their search limits due to limited observability and dataset coupling. By strictly enforcing physically motivated bounds, the optimization successfully prevented these variables from artificially absorbing unmodeled mechanical backlash. This avoided physically implausible skew configurations and ultimately delivered highly robust out-of-sample generalization.

\section{Conclusion and Future Work}

This paper introduced a compact, robust bound-constrained kinematic calibration framework for a 6-DoF serial manipulator, tailored specifically for the strict precision requirements of robotic needle guidance. While conventional calibration pipelines typically treat parameter bounds merely as numerical safeguards, the proposed method successfully demonstrated that physically motivated bounds can act as an explicit regularization mechanism. By combining observability-optimized sampling, Kabsch initialization, and bounded Trust-Region Reflective optimization, the framework effectively mitigated parameter coupling and overfitting.

Extensive evaluation across simulation, 10-fold cross-validation, and independent hardware experiments confirmed that this constrained geometric approach consistently delivers robust, sub-millimeter positioning accuracy. These outcomes validate that treating physical constraints as active design elements yields a stable calibration model, establishing the necessary kinematic foundation for advanced percutaneous interventions.

\subsection{Future Work}
\label{subsec:future_work}

Several directions can further strengthen the proposed framework. First, the current model is limited to geometric errors and does not explicitly capture non-geometric effects such as joint compliance, backlash, gravitational deflection, or thermal drift, which likely account for the residual gap between simulated and physical performance. Extending the identification framework to include these effects, or combining the geometric calibration with lightweight data-driven residual compensation, could further improve absolute positioning accuracy.

Additionally, more extensive validation across larger pose sets, particularly near workspace boundaries and joint limits, would strengthen confidence in global generalization. Furthermore, to definitively evaluate the efficacy of the bound-constrained approach, future studies will incorporate direct empirical benchmarking against other state-of-the-art kinematic calibration algorithms. Finally, translation toward clinically relevant applications, such as percutaneous thermal ablation, will require validation beyond laboratory tracking. Future efforts will focus on phantom studies and \textit{ex vivo} soft-tissue experiments under realistic constraints, including simulated respiratory motion and procedure-specific collision boundaries.

\section*{Acknowledgment}

The authors would like to thank Precision Medtech Co., Ltd., and the Pattern Recognition Lab at Friedrich-Alexander-Universität Erlangen-Nürnberg for their support with the robotic infrastructure and lab environment.




\begin{sloppypar}
\bibliographystyle{IEEEtran}
\bibliography{ref.bib}
\end{sloppypar}

\vspace{12pt}
\color{red}

\end{document}